# Comparison between EM and FCM algorithms in skin tone extraction

*Elham Ravanbakhsh*
dep. of engineering, Shahid chamran university of Ahvaz
Ahvaz, Iran
e-ravanbakhsh@stu.scu.ac.ir

*Ehsan Namjoo*
dep. of engineering, Shahid chamran university of Ahvaz
Ahvaz, Iran
e.namjoo@scu.ac.ir

*Mosab Rezaei*
dep. of engineering, Shahid chamran university of Ahvaz
Ahvaz, Iran
m-rezaei@stu.scu.ac.ir

*Padideh Choobdar*
dep. of engineering, Shahid chamran university of Ahvaz
Ahvaz, Iran
p-chubdar@stu.scu.ac.ir

*Abstract*— this study aims to investigate implementing EM and FCM algorithm for skin color extraction. The capabilities of three well-known color spaces, namely, RGB, HSV and YCbCr for skin-tone extraction are assessed by using statistical modeling of skin-tones using EM and FCM algorithms. The results show that utilizing a Gaussian mixture model for parametric modeling of skin-tones using EM algorithm works well in HSV color space when all three components of color vector are used. In spite of discarding the luminance components in YCbCr and HSV color spaces, EM algorithm provides the best results. The results of detailed comparison are explained in conclusion.

*Keywords— EM; FCM; ROC; skin color segmentation; SPM*

## I. Introduction

Skin region extraction in color images could be the first step for face detection algorithms in color images. In [1] some methods including LUT[1], SOP[2] and Gaussian mixture model are used for skin color modeling. In [1] it has been claimed that methods based on Bayesian model and maximum entropy model have best results in RGB color space. In [2] two different methods have been used to extract skin regions. The first method uses mixture factor analyzer to cluster skin samples, Then PCA is used to reduce dimensionality and, finally model parameters are estimated by EM algorithm. In the second method, Kohonen's Self Organizing Map is used to classify skin samples, then Fisher Linear Discriminant analysis method is used to reduce dimensionality and finally Maximum Likelihood Estimation estimates the parameters of the model.
In [3] the performance of K-means and Connected Component Algorithm to extract skin regions is Compared. Compared to K-means algorithm, the results represent efficiency of Connected Component Algorithm. In [4] in order to extract skin regions, range of each component of color space model is explicitly defined. After extracting skin regions, the authors have used morphological operators to propose a method to recognize faces in color images.
One of effective and efficient algorithms which is widely used in pattern recognition, is Expectation Maximization (EM) algorithm. It can be considered as a powerful unsupervised learning method to be used in clustering algorithms. In [5] this method is used to extract skin region. Another effective method widely used for clustering application, is Fuzzy C-Means (FCM) algorithm[6]. This method in [7] is used to extract skin regions too.
Applying an effective clustering method plays an important role in the success of skin region segmentation. This paper tries to evaluate the clustering performance of two popular and conventional algorithms, EM and FCM in different color space models. Comparison is made according to Receiver Characteristic Operating (ROC) curves. On the other hand, according to results, the capability of each color space for skin-tone modeling is also evaluated. More details will be discussed in next sections.
The remainder of this paper is organized as follows. In sections 2 and 3, a brief mathematical background including Gaussian mixture model and FCM clustering algorithm is presented. Section 4 is devoted to the main goal of the paper, i.e., the detailed comparison of clustering algorithms and color spaces. Finally, in section 5, the paper will be concluded.

## II. Fuzzy C-Means

FCM is one of the most popular clustering algorithms widely used to categorize input samples based on their similarities. In this method, the samples are assigned to *C* clusters. Associated with each sample, there is a set of membership levels that indicates the probabilities that the sample is belonged to any of the *C* predefined clusters. FCM tries to minimize the objective function that can be defined for n sample as equation (2-1).

---

[1] Normalized Lookup

[2] Self-Organized Map

$$L = \sum_{i=1}^{C}\sum_{j=1}^{n} \delta_{ij}{}^m \|x_j - \mu_i\|^2 \qquad (2\text{-}1)$$

In (2-1) m is fuzzifier and determines the level of the clusters' fuzziness. The value of this parameter is chosen to be a real number greater than one. In this paper, the value of this parameter is set to 2. Also $\delta_{ij}$ determines the degree to which sample $x_i$ is belonged to cluster $C_j$. $\mu_i$ is the mean of cluster $i$ and $x_j$ represents $j^{th}$ sample. $\|x_j - \mu_i\|^2$ is the Euclidian distance between $j^{th}$ sample and $i^{th}$ cluster center.

Membership value of a sample in all clusters is normalized as below.

$$\sum_{i=1}^{C} \delta_{ij} = 1 \qquad j = 1,\ldots,n \qquad (2\text{-}2)$$

To minimize the objective function, one can set $\frac{\partial L}{\partial \mu_i} = 0$ and $\frac{\partial L}{\partial \delta_{ij}} = 0$, then optimum values of $\delta_j$ and $\mu_i$ is achieved as follows.

$$\mu_i = \frac{\sum_{i=1}^{n} \delta_{ij}^m \cdot x_i}{\sum_{i=1}^{n} \delta_{ij}^m} \qquad (2\text{-}3)$$

And

$$\delta_{ij} = \frac{1}{\sum_{k=1}^{C}\left(\frac{\|x_j - \mu_i\|}{\|x_k - \mu_i\|}\right)^{\frac{1}{m-1}}} \qquad (2\text{-}4)$$

The pseudo-code of FCM algorithm is as follows.

Algorithm FCM:
1. Begin
2. initialize $x_1, \ldots x_n$, C, m, ε, max_iteration
3. generate randomly $\mu_i$
4. compute $\delta_{ij}$ by Eq. (2-4)
5. do{
6. compute $\mu_i$ by Eq. (2-3)
7. compute $\delta_{ij}$ by Eq. (2-4) }
8. while (change in $\delta_{ij}$ is more than ε) and (number of iterations is less than max_iteration)
9. return $\mu_i, \delta_{ij}$
10. End

At last, sample $x_i$ belongs to cluster $j$ if:

$$\delta_{ij} > \delta_{ij\prime} \qquad j\prime \neq j \quad \text{for all j'}$$

## III. GASSIAN MIXTURE MODEL(GMM)

For n independent observed samples $\{y_1, y_2, \ldots, y_n\}$, their actual probability density function could be estimated as a Gaussian mixture model like equation (3-1).

$$f(y_t, \psi) = \sum_{p=1}^{P} \pi_p f(y_t, \theta_p) \qquad (3\text{-}1)$$

$\psi$ is the parameter vector of Gaussian mixture model and it is equal to $(\pi_1, \pi_2, \ldots \pi_{p-1}, \theta_1, \theta_2, \ldots \theta_p)$. $\theta_i$ s for i =1,2,…,p are the parameter vector corresponding to function $f(y_t, \theta_p)$ and $\pi_i$'s are prior probability ($\sum_i \pi_i = 1$).

There are several approaches to estimate GMM's parameters. One of the most successful and effective approaches for estimation is maximum likelihood estimation (MLE). Expectation Maximization (EM) algorithm uses MLE and Bayes decision rule to iteratively find these parameters. The detailed description of EM algorithm can be found in [8].

## IV. SIMULATION RESULTS

In this section, the performance of clustering algorithms in skin region extraction is compared. To this aim, the *university of Tabriz* face dataset is used. The dataset is available in [9]. 440 images from the dataset have been selected and total number of 23000 skin-tone pixels have been extracted out of these images to compose the training skin-tone set. Performance evaluation is organized as follows. Images are selected in different lightening conditions. Also the resolution of images are different and mostly in 2816*2112 and 640*480 resolution.

A Gaussian mixture model is considered as the probability density function of real skin samples in different color spaces. The parameters of the mixture for each color space are estimated once using EM algorithm and once using FCM. To make the comparison fair, the number of clusters is considered the same and equal to 3 for both algorithms.

In this work three well-known color spaces, i.e. RGB, YCbCr and HSV are used. To evaluate luminance effect, in YCbCr and HSV, simulations have been done once for all components of color space model and once again just for choroma components, i.e. Cb and Cr from YCbCr, and H and V from HSV. For RGB color space model and for all cases, simulations have been done for normalized components R and G as shown in equations (4-1) and (4-2).

r = R/R+G+B (4-1)

g = G/ R+G+B (4-2)

Skin color modeling using EM algorithm in YCbCr is managed as follows. First, RGB samples are mapped into YCbCr color space; afterwards, a Gaussian mixture model like (4-3) is fitted to samples.

$$f(C|skin) = \sum_{i=1}^{3} \pi_i f_x(\theta_i) \qquad (4\text{-}3)$$

In equation (4-3), $\pi_i$ is prior probability of cluster i and $\theta_i$ is the parameter vector of cluster *i* including mean vector ($\mu_i$) and covariance matrix ($\Sigma_i$) and *C* is the color vector. For example in YCbCr color space C includes Y, Cb and Cr, and in the case of luminance elimination, it would include just Cb and Cr.

In the next step, for each pixel from a test image, *f(C|skin)* is calculated and its value is replaced. Applying this procedure to all other pixels causes a grey-level image called Skin Probability Map (SPM). Creating SPM in RGB and HSV model is the same. The SPM could be created just using two components from color vector *C*.

Skin color modeling using FCM in YCbCr is similar to what is done using EM. The only difference is that the parameters of GMM are achieved using FCM algorithm. To this aim, when the clustering algorithm using FCM is finished, then for $i^{th}$ cluster, the mean vectors ($\mu_i$) and covariance matrix ($\Sigma_i$) are calculated. Furthermore the prior ($\pi_i$) for class *i* is calculated as follows:

$\pi_i$ = (number of samples in cluster i) / (total number of samples)

In parametric methods, a reliable metric is needed to evaluate performance of the model. In this work ROC curves is used for evaluating the performance of GMMs.

In an SPM image, the values of all pixels is normalized into interval [0,1], so pixels with values close to 1 are likely to be actual skin pixel in corresponding color image. By thresholding SPM image with some value in this interval, a binary image is created. In this case *True positive(TP)* is the number of pixels with value equal to 1 in the binary image that are in actuality corresponded to real skin pixels in related original test image. Similarly *False Positive (FP)* is the number of pixels with value equal to 1 in binary image that are not in actuality corresponded to real skin pixels in related original test image. True Positive Rate and False Positive Rate are also defined as follows.

TPR=TP/S  (4-4)

FPR=FP/NS  (4-5)

S is the total number of skin pixels and NS is total number of non-skin pixels. The scenario has to be followed to create a ROC curve, is presented in the next paragraph.

Consider a sequence of K sorted values from some small value close to 1, say 0.001, to some value close to 1, say, 0.999 by a small step size like 0.001. For the k'th entity in the sequence, one can create a binary image by thresholding the SPM image with this entity. So for each value from the defined sequence, there would be a sequence of *K* ponits *{(TPR_k,FPR_k), for k = 1,2,...,K}* in TPR/FPR plane. Connecting these points, a ROC curve is created.

In each picture in addition to real skin regions, there are some regions similar to actual skin tones, existence of these regions make the skin color extraction hard. In following the performance of algorithms in discriminating skin-tone pixels will be assessed using ROC curves. Figure (1) shows a sample test image from the dataset and Figure (2) shows its corresponding SPM.

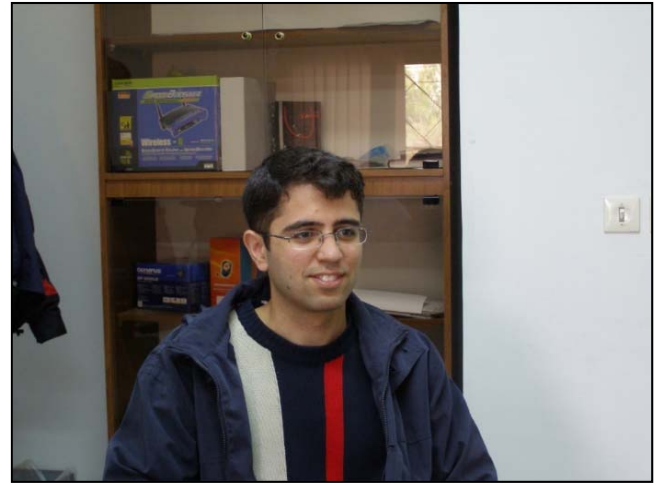

Fig. 1. A sample of test image

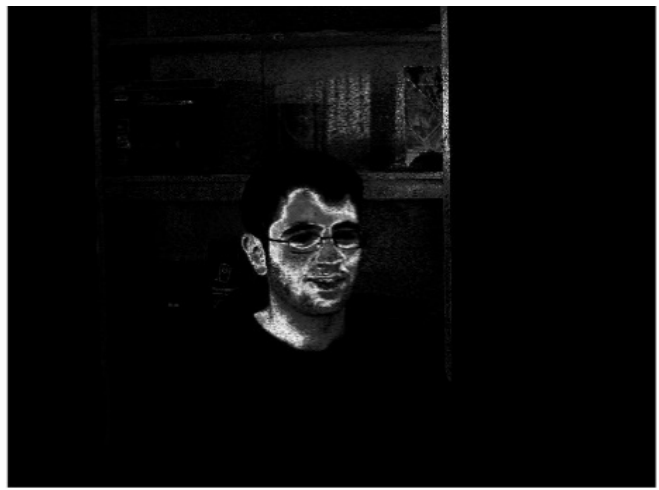

Fig. 2. SPM image corresponded to figure (1)

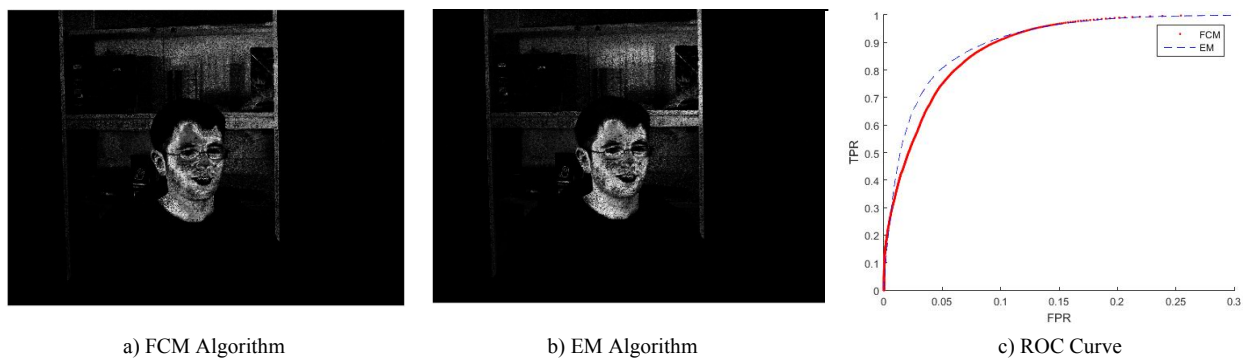

a) FCM Algorithm  b) EM Algorithm  c) ROC Curve

Fig. 3. SPMs and ROCs by FCM and EM algorithm with 3 components in RGB color space

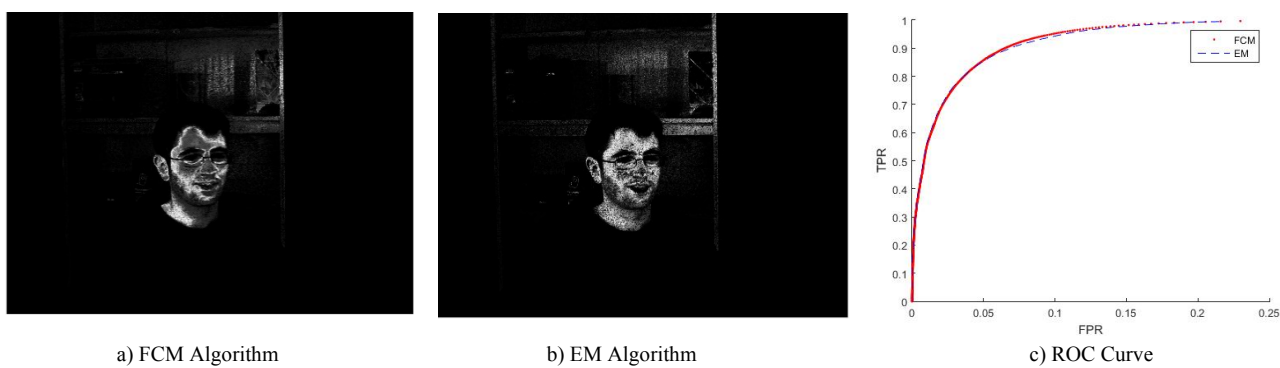

a) FCM Algorithm  b) EM Algorithm  c) ROC Curve

Fig. 4. SPMs and ROCs by FCM and EM algorithm with 3 components in HSV color space

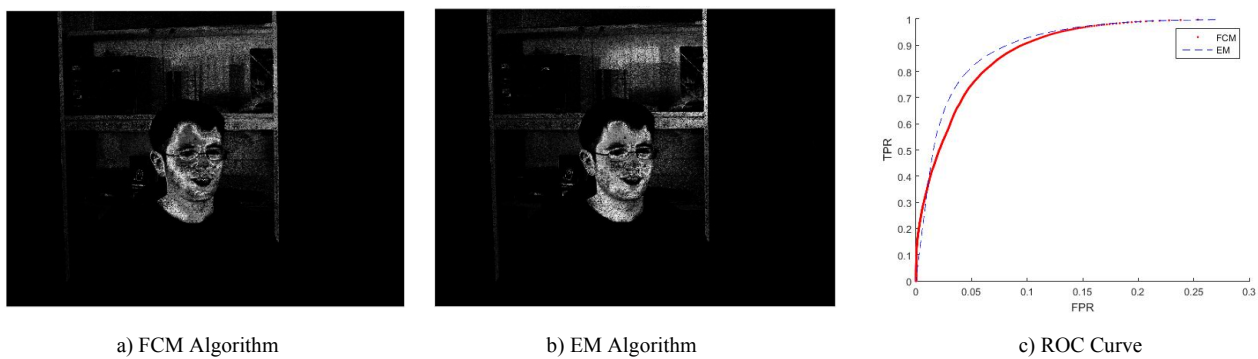

a) FCM Algorithm  b) EM Algorithm  c) ROC Curve

Fig. 5. SPMs and ROCs by FCM and EM algorithm with 3 components in YCbCr color space

According to figures (3), (4) and (5) it can be concluded that when all color components are used, EM algorithm has better results rather than FCM in RGB and YCbCr color spaces. But in HSV there is not any big difference and both algorithms are approximately equivalent.

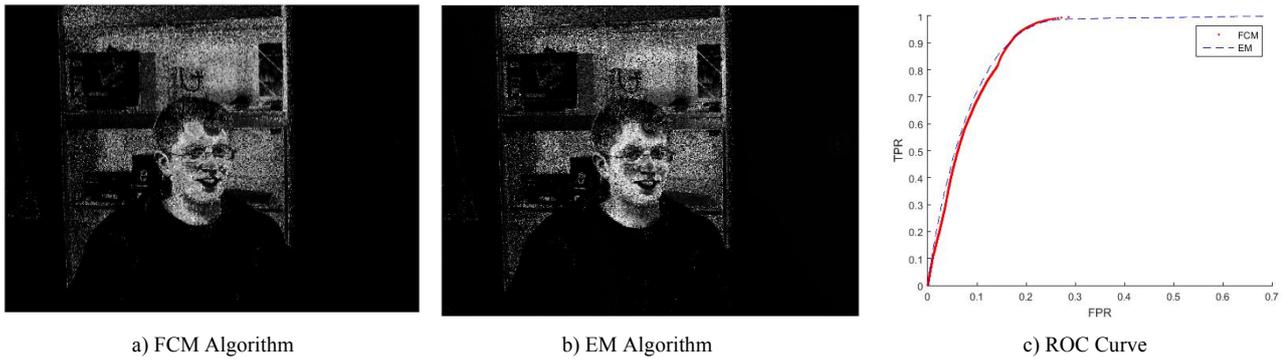

a) FCM Algorithm     b) EM Algorithm     c) ROC Curve

Fig. 6. SPMs and ROCs by FCM and EM algorithm with 2 components in RGB color space

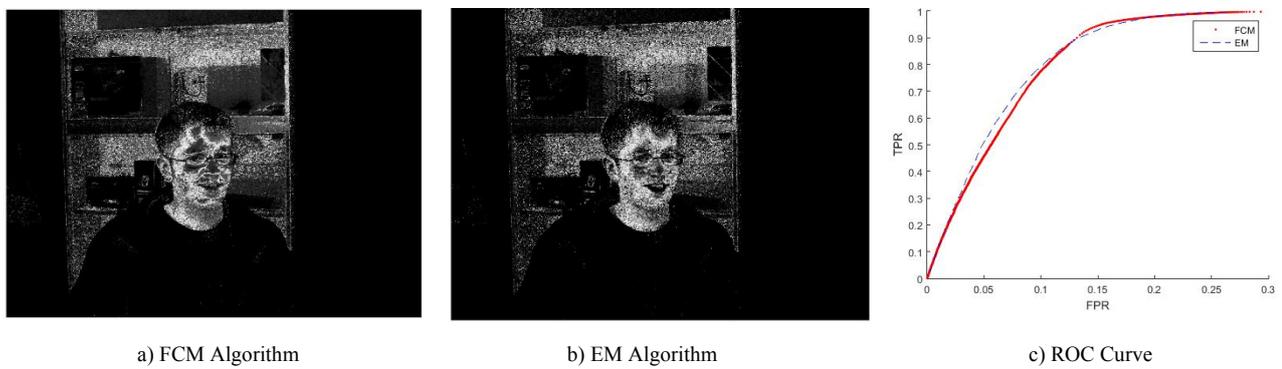

a) FCM Algorithm     b) EM Algorithm     c) ROC Curve

Fig. 7. SPMs and ROCs by FCM and EM algorithm with 2 components in HSV color space

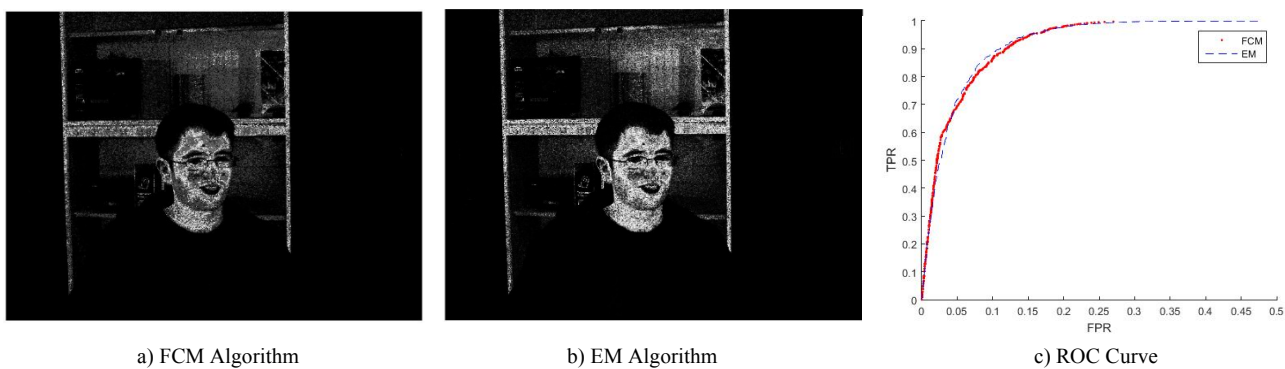

a) FCM Algorithm     b) EM Algorithm     c) ROC Curve

Fig. 8. SPMs and ROCs by FCM and EM algorithm with 2 components in YCbCr color space

As figures (6), (7) and (8) illustrate. One can also conclude when luminance elimination is done for HSV and YCbCr color spaces, results of FCM and EM algorithms do not differ in a big deal. Also there is no significant difference between EM and FCM with eliminating B component in RGB color space.

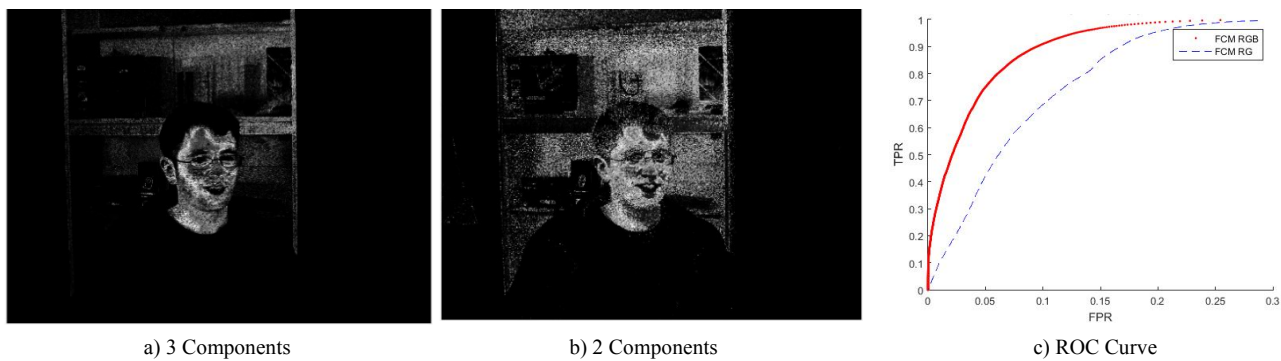

a) 3 Components     b) 2 Components     c) ROC Curve

Fig. 9. SPMs and ROCs by FCM with 3 and 2 components in RGB color space

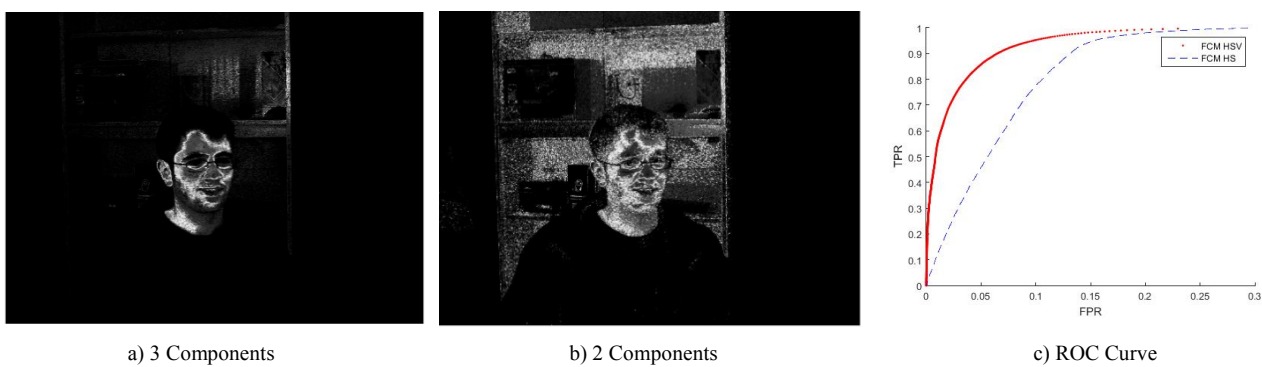

a) 3 Components     b) 2 Components     c) ROC Curve

Fig. 10. SPMs and ROCs by FCM with 3 and 2 components in HSV color space

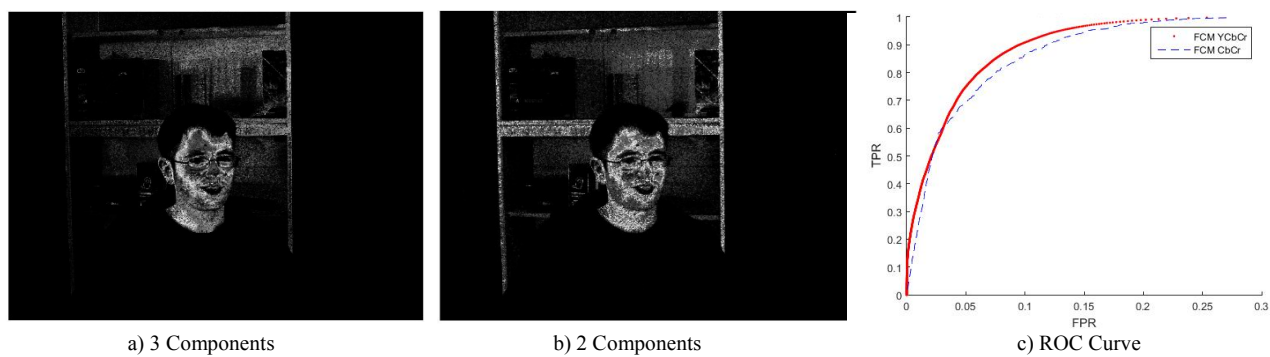

a) 3 Components     b) 2 Components     c) ROC Curve

Fig. 11. SPMs and ROCs by FCM with 3 and 2 components in YCbCr color space

According to figures (9), (10) and (11), by applying FCM, the ROC curves show that using all three components have better results rather than 2 when just two components are used. But when YCbCr color space is used, there is no sensible difference between choosing two or three color components.

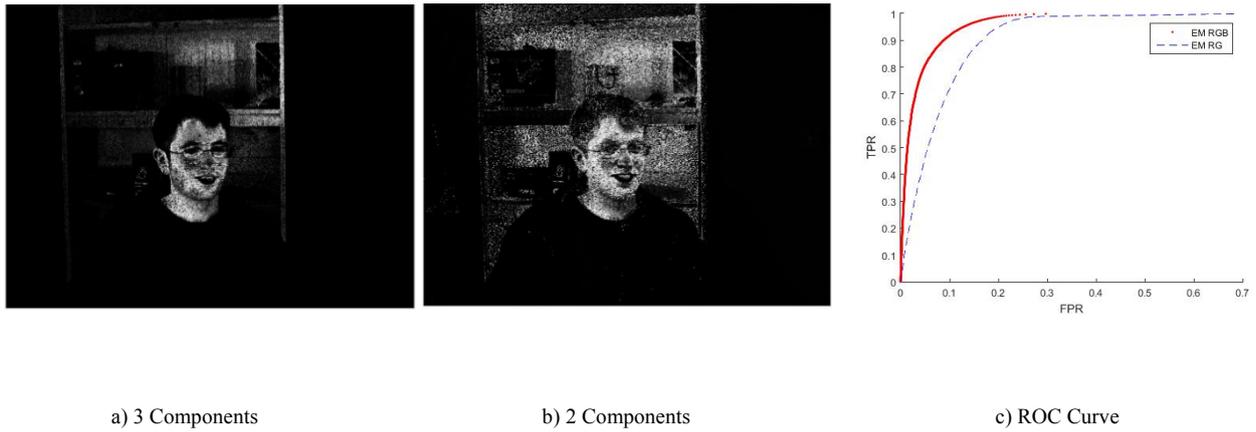

a) 3 Components  b) 2 Components  c) ROC Curve

Fig. 12. SPMs and ROCs by EM with 3 and 2 components in RGB color space

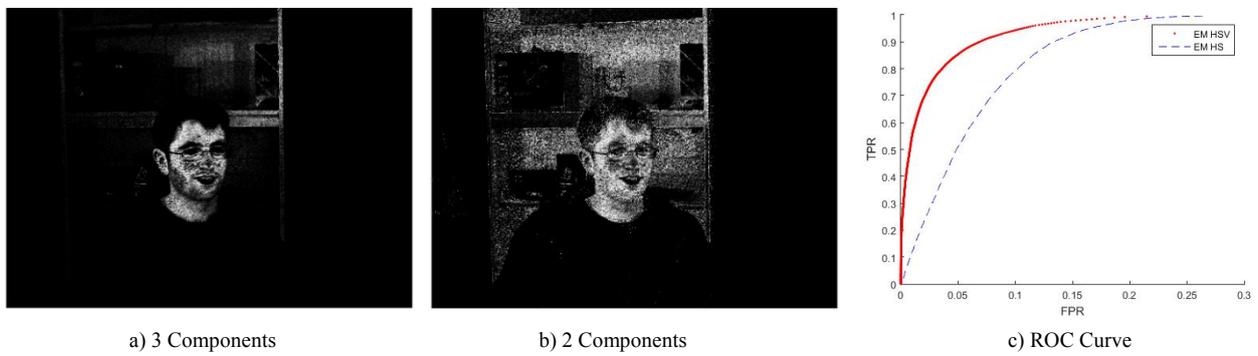

a) 3 Components  b) 2 Components  c) ROC Curve

Fig. 13. SPMs and ROCs by EM with 3 and 2 components in HSV color space

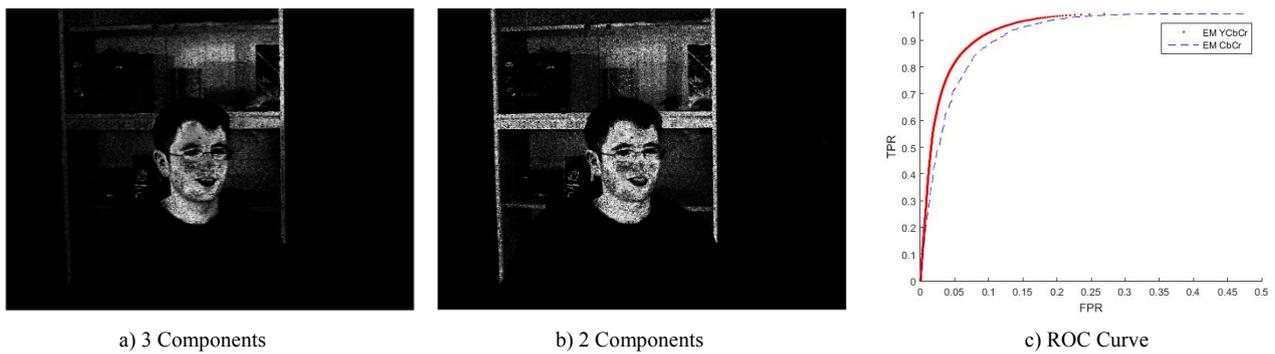

a) 3 Components  b) 2 Components  c) ROC Curve

Fig. 14. SPMs and ROCs by EM with 3 and 2 components in YCbCr color space

Also, According to figure (12), (13) and (14), by applying EM, the ROC curves show that using three components of color space in RGB and HSV have much better results rather than when just two components are used. But when YCbCr color space is used, there is no sensible difference between choosing two or three color components.

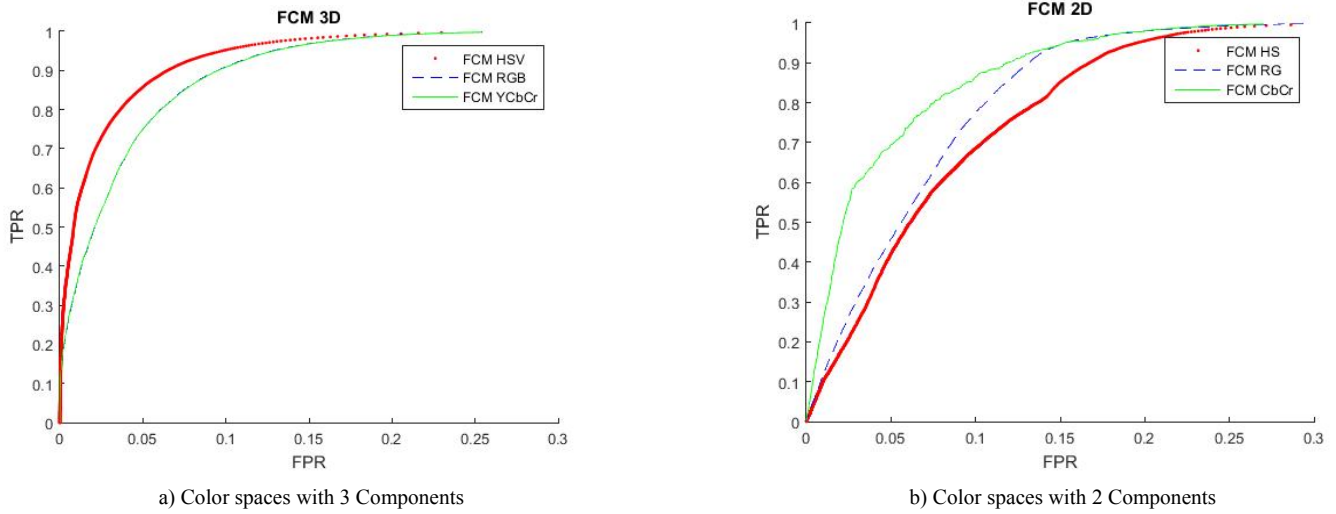

a) Color spaces with 3 Components      b) Color spaces with 2 Components

Fig. 15. ROCs by FCM with 3 and 2 components in 3 color spaces

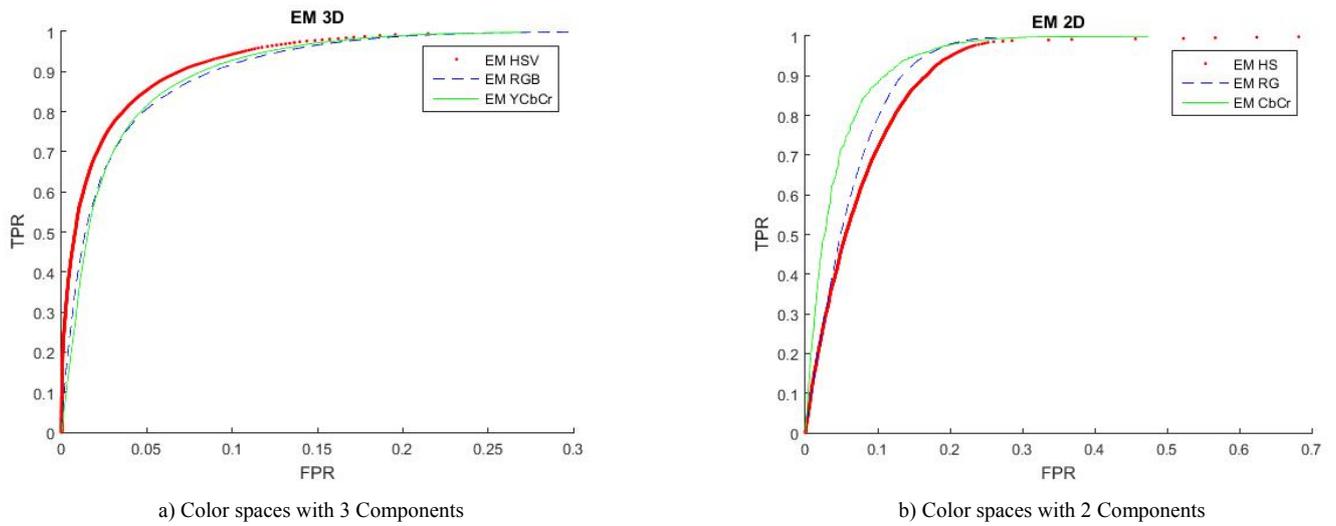

a) Color spaces with 3 Components      b) Color spaces with 2 Components

Fig. 16. ROCs by EM with 3 and 2 components in 3 color spaces

Finally, in figure (15) and (16) the performances of EM and FCM algorithms have been compared using both three and two components, in three color spaces. Results illustrate that when all color components are used, HSV model has better results rather than the others. It can also be concluded when a component of color space is eliminated, YCbCr color space results is much more robust in comparison to the others. Since the eliminated component in HSV and YCbCr is luminance, it can be concluded that YCbCr color space is much more robust against luminance eliminating in comparison with HSV color model.

V. CONCLUSION

According to the ROC curves, we can conclude that EM algorithm is more successful than FCM algorithm in skin region extraction, especially in RGB and YCbCr color spaces. Also it was observed that eliminating the luminance component, in this case; do not affect the results drastically. Similarly, it was shown that in both algorithms when luminance component is discarded, the accuracy is decreased; except for YCbCr color space where in, both algorithms have similar result with or without discarding luminance component.

In general, we can conclude that best results are obtained in HSV color space without luminance elimination, and best results by eliminating luminance component is obtained when YCbCr color space is used.

To finish, we can conclude that firstly, when all components of color space are used, EM algorithm in HSV color space provides the best results in comparison with FCM algorithm for skin region extraction. Secondly, when discarding luminance component is desired, both algorithm could achieve similar results in YCbCr color space and there is no vivid different among them.